\pdfoutput=1

\documentclass[11pt]{article}
\usepackage[]{acl}

\usepackage[T1]{fontenc}

\usepackage{algorithm}
\usepackage{algorithmic}
\usepackage{times}
\usepackage{latexsym}
\usepackage[utf8]{inputenc}
\usepackage{microtype}
\usepackage{xspace}
\usepackage{caption}
\usepackage{subcaption}
\usepackage{amsmath}
\usepackage{url}
\usepackage{amssymb}
\usepackage{bbding}
\usepackage{amsfonts}
\usepackage{graphicx}
\usepackage{tabularx}
\usepackage{multirow, multicol}
\usepackage{arydshln}
\usepackage{mathtools,nccmath}
\usepackage{enumitem}
\usepackage{todonotes}
\usepackage{cleveref}
\usepackage{booktabs}
\usepackage{CJKutf8}
\usepackage{color}
\usepackage{hyperref}

\definecolor{mydarkgreen}{RGB}{0, 139, 69}

\newcommand{\ti}[1]{\textit{#1}}
\newcommand{\tif}[1]{\textit{\textbf{#1}}}

\newcommand{\eg}{e.g.\@\xspace}

\newcommand{\ourmodel}{\textsc{UniSAr}\xspace}

\title{\textsc{UniSAr}: A Unified Structure-Aware Autoregressive Language Model for Text-to-SQL}

\author{ Longxu Dou$^{1}$\thanks{~ Contribution during the internship at Microsoft Research Asia.
	}, Yan Gao$^{2}$, Mingyang Pan$^{1}$, Dingzirui Wang$^{1}$, \\ \textbf{Wanxiang Che$^{1}$, Dechen Zhan$^{1}$, Jian-Guang Lou$^{2}$} \\
	$^{1}$    Harbin Institute of Technology,\\
	$^{2}$    Microsoft Research Asia\\
	{\tt \{lxdou, mypan, dzrwang, car\}@ir.hit.edu.cn}, {\tt dechen@hit.edu.cn}    \\
	{\tt \{Yan.Gao, jlou\}@microsoft.com}    \\
}

\begin{document}
	\maketitle
	\begin{abstract}
		Existing text-to-SQL semantic parsers are typically designed for particular settings such as handling queries that span multiple tables, domains or turns which makes them ineffective when applied to different settings.
		We present \ourmodel (\tif{Uni}fied \tif{S}tructure-\tif{A}ware Auto\tif{r}egressive Language Model), which benefits from directly using an off-the-shelf language model architecture and demonstrates consistently high performance under different settings.
		Specifically, \ourmodel extends existing autoregressive language models to incorporate three non-invasive extensions to make them \ti{structure-aware}:
		(1) adding \ti{structure mark} to encode database schema, conversation context, and their relationships; 
		(2) \ti{constrained decoding} to decode well structured SQL for a given database schema;
		and (3) \ti{SQL completion} 
		to complete potential missing \textsc{JOIN} relationships in SQL based on database schema.
		On seven well-known text-to-SQL datasets covering multi-domain, multi-table and multi-turn, \ourmodel demonstrates 
		highly comparable or better
		performance to the most advanced specifically-designed text-to-SQL models.
		Importantly, our \ourmodel is non-invasive, such that other core model advances in text-to-SQL can also adopt our extensions to further enhance performance
		\footnote{Codes and checkpoints are available at \href{https://github.com/microsoft/ContextualSP/tree/master/unified_parser_text_to_sql}{link}.}.
		
	\end{abstract}
	
	\section{Introduction}\label{sec:introduction}
\begin{figure} 
		\centering
		\includegraphics[width=0.9\linewidth]{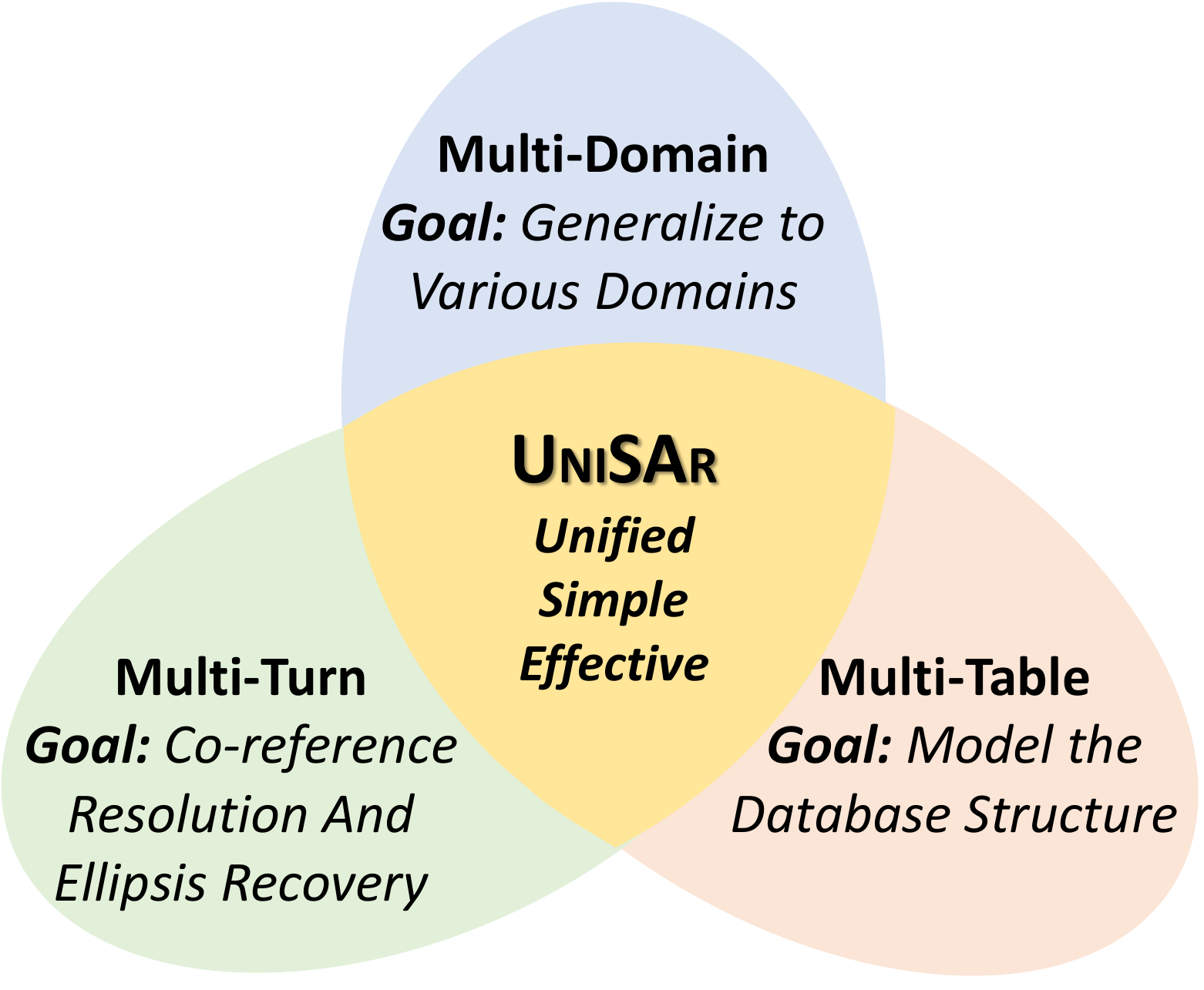}
		\caption{
			Our view of three main research settings in text-to-SQL.
		}
		\label{fig:framework_comparsion}
	\end{figure}

	Text-to-SQL translates a user's natural language question into a corresponding SQL query
	\cite{zhongSeq2SQL2017,yu-etal-2018-spider,guo-etal-2019-towards,yu-etal-2019-sparc}. This greatly reduces the entry barrier of data analysis to lay users daunted by the technical nuances of SQL.
	As text-to-SQL techniques matured, enhancements have been proposed to tackle different settings. 
	These enhancements can be roughly organized into three settings of research (Figure~\ref{fig:framework_comparsion}):
	(1) \textit{multi-domain} where a parser must generalize to databases in various domains \cite{zhongSeq2SQL2017, sun2020tableqa};
	(2) \textit{multi-table} where the parser must understand the database structure and generate complex SQL query bridging multiple tables \cite{yu-etal-2018-spider, wang-etal-2020-dusql}; and 
	(3) \textit{multi-turn} where a parser must understand the dialog history, often requiring co-reference resolution and ellipsis recovery~\cite{yu-etal-2019-cosql, yu-etal-2019-sparc, guo-etal-2021-chase}.
	
	To address the unique challenges of different settings, researchers have proposed various neural architectures, such as grammar-based decoder for SQL generation~\cite{yin-neubig-2018-tranx,guo-etal-2019-towards,wang-etal-2020-rat}, GNN for database structure modeling~\cite{bogin-etal-2019-global}, and stacked interaction-layer for context modeling~\cite{cai-wan-2020-igsql}.
	However, these architectures are prone to `overfitting' specific datasets, making them non-trivial to adapt to others. For example, the parsers developed for the WikiSQL~\citep{zhongSeq2SQL2017} typically cannot work well on Spider benchmark~\citep{yu-etal-2018-spider}.
	
	In this work, we present a \tif{simple} yet \tif{effective} text-to-SQL parser: \ourmodel (\tif{Uni}fied \tif{S}tructure-\tif{A}ware Auto\tif{r}egressive Language Model).
	Compared with the specifically-designed models (invasive), \ourmodel is \tif{simple} as it does not need any specifically designed DNN modules other than a pre-trained language model (non-invasive).
	Such a simple architecture gives \ourmodel the opportunity to be easily adapted to different datasets, and thus \ourmodel enjoys high \tif{generalizability}.
	Besides, benefiting from our proposed non-invasive extensions, \ourmodel achieves consistently high performance across different settings, which is very \tif{effective}.
	Concretely, we propose three non-invasive extensions to make an off-the-shelf autoregressive language model structure-aware.
	First, we encode structural information (\eg{ database schema, conversation context and their relationships}) by inserting some special tokens named \ti{structure marks} into the serialized schema and question as inputs. 
	Second, we adopt \ti{constrained decoding} to avoid the decoder generating invalid tokens (e.g., synonyms of schema) during SQL generation following \citet{Scholak2021:PICARD, decao2020autoregressive, decao2020multilingual}.
	Finally, we propose \ti{SQL completion} to make the SQL complete through inferring potential missing \textsc{JOIN} components based on the database schema.

	To prove the effectiveness and generalizability of our \ourmodel, we conduct experiments on seven popular text-to-SQL datasets covering multi-domain, multi-table and multi-turn.
	With a simple and unified architecture, our model achieves comparable or even better performance against task specific models. 
	Importantly, the simple architecture enables different tasks to share the sample training protocol. 
	\ourmodel could be easily improved by multi-task training.
	
	To summarize, our contributions are three-fold:
	\begin{itemize}
		\item We propose a unified parsing framework for various text-to-SQL settings including
		multi-domain, multi-table and multi-turn, without reliance on specific architecture designs.
		\item We make autoregressive language models structure-aware via simply incorporating three non-invasive extensions: structure mark, constrained decoding and SQL completion. 
		\item We conduct extensive experiments on seven text-to-SQL datasets to demonstrate the better effectiveness of our unified parser \ourmodel compared with specifically-designed baselines. 

		and achieves overall improvements. 
	\end{itemize}

	\begin{figure*} 
		\centering
		\includegraphics[width=1\linewidth]{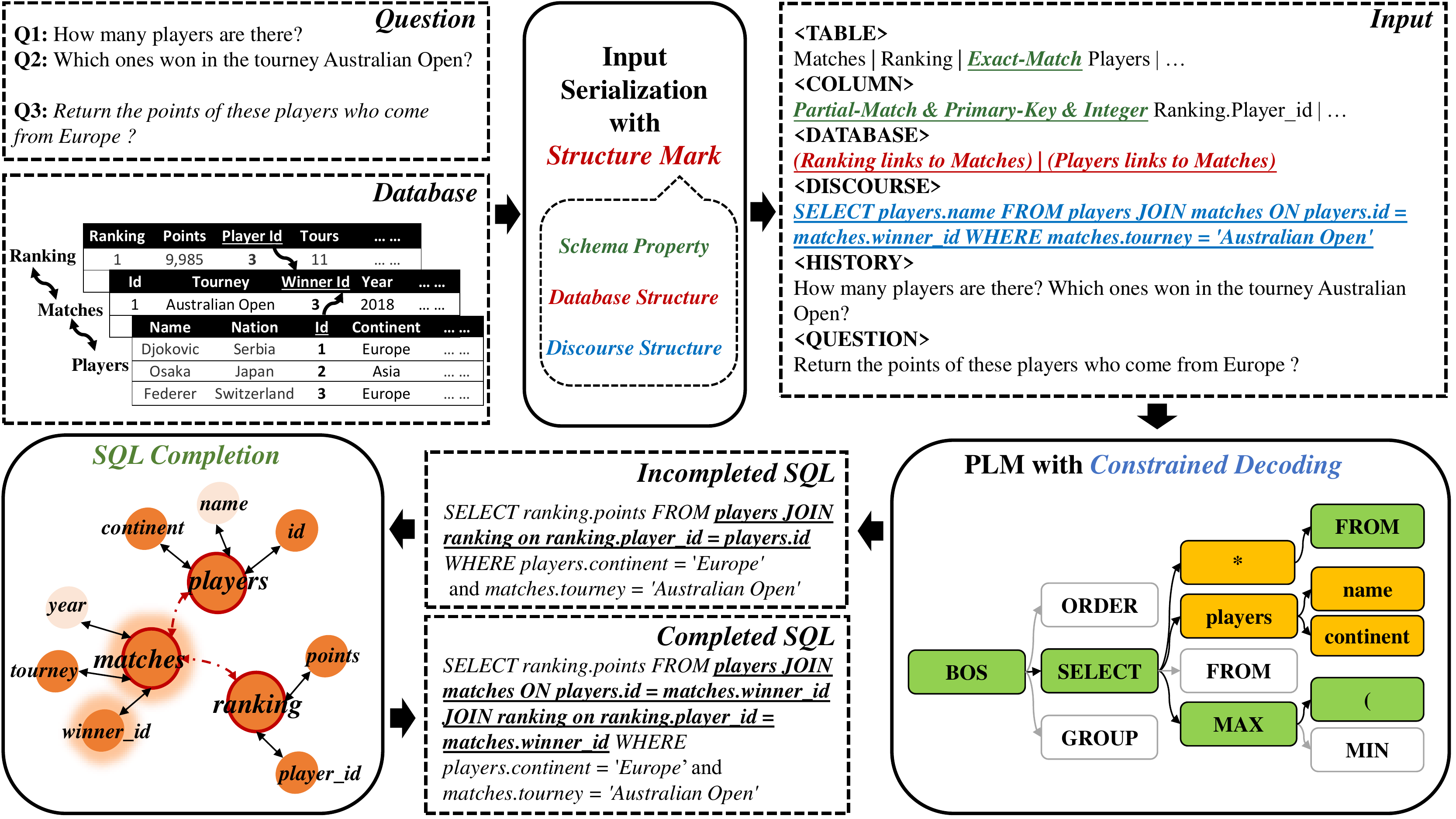}
		\caption{
			Three non-invasive extensions to make PLMs become structure-aware: 
			\textit{(1)} add \tif{structure mark} to encode database schema, conversation context and their relationships; 
			\textit{(2)} \tif{constrained decoding} to decode well-structured SQL;
			\textit{(3)} \tif{SQL completion} through inferring the potential missing \textsc{JOIN} components based on the database schema.
		}
		\label{fig:mark_pipeline}
	\end{figure*}

	\section{Related Work}
	\paragraph{Invasive Approaches}
	Overall, the recent state-of-the-art models for text-to-SQL use various specific architectures for question/schema encoding and SQL query decoding.
	Take some popular models for example. 
	To joint encode the question and schema, 
	\citet{xu2017sqlnet} proposed column attention strategy to gather information from columns for each question word. 
	EditSQL~\citep{zhang-etal-2019-editing} considered co-attention between question words and database schema nodes. 
	\citet{bogin-etal-2019-representing} dealed with the graph structure of database schema via GNN.
	RAT-SQL~\citep{wang-etal-2020-rat}, utilized a complete relational graph attention neural network to handle various pre-defined relations.
	To ensure the syntactic and semantic correctness of the generated SQL query, existing works~\citep{guo-etal-2019-towards,wang-etal-2020-rat} usually adopted grammar-based decoder.
	In contrast, our \ourmodel provides a unified way to encode structural information and decode valid SQL.
	It is very simple as it does not need specific designed modules (non-invasive) and also incredibly effective like these invasive approaches.
	
	\paragraph{Treat Text-to-SQL as Seq2Seq Task}
	We use the same task formulation as \citep{dong-lapata-2016-language,zhongSeq2SQL2017,lin-etal-2018-nl2bash,2020t5} that treated text-to-SQL as a translation problem. 
	However, they only solve it by applying seq2seq models on sequence of tokens.
	Thus they can't productively encode the input structure.
	We extend their work with three non-invasive extensions
	to effectively encode the input structure.
	To the best of our knowledge, this is the first time that seq2seq parsers are on par with specially-designed parsers 
	in various text-to-SQL settings. 
	
	\paragraph{Non-invasive Approaches}
	Recently, \citet{Scholak2021:PICARD} proposed PICARD, a similar approach to utilizing language models for text-to-SQL, which focus on constraining auto-regressive decoders of language models through incremental parsing.
	The main difference between PICARD and \ourmodel is that they only considered the structure in the decoding phrase but ignored the input structure.
	Our proposed extensions like structure mark could further improve PICARD model (Table~\ref{tab:compare_with_other_PLM}).
	Moreover, \ourmodel is basically built on moderately-sized language model BART-Large (400M). 
	

	\section{Methodology}
	
	Overall, we employ pretrained autoregressive language models as the backbone of \ourmodel. 
	since they exhibit excellent adaptability and generalizability in many NLP tasks.
	As shown in Figure~\ref{fig:mark_pipeline}, we propose three non-invasive extensions to make PLMs become structure-aware:
	\textit{(1)} we encode structural information (\eg{ database schema, conversation context and their linking relationships}) by inserting some special tokens named \ti{structure marks} into the serialized schema and question as inputs; 
	\textit{(2)} we adopt \ti{constrained decoding} to decode well-structured SQL via simply filtering invalid tokens (e.g., synonyms of schema) during beam search.
	\textit{(3)} we propose \ti{SQL completion} to infer the underlying \textsc{JOIN} relationships in the SQL statement based on the predicted incomplete SQL and the database schema.


	\subsection{Pretrained Language Model}
	In our experiments, we implement \ourmodel on top of BART~\citep{lewis-etal-2020-bart}, a widely used pre-trained encoder-decoder model.
	BART follows a standard sequence-to-sequence Transformer architecture \cite{Vaswani2017AttentionIA}.
	It is pre-trained via corrupting sentences (e.g., randomly sampling spans and masking each one 
	) and then optimizing a reconstruction loss. 
	The reason using BART not T5 here is that it's show excellent copy ability according to preliminary experiments\footnote{We employ the BART-Large for English and the mBART-CC25 for Chinese. For brevity, we collectively call them BART in tables.}.

	\subsection{Formulate Text-to-SQL as Seq2Seq}
	The input for \ourmodel contains a series of NL questions and database.
	We give an example of multi-turn in Figure~\ref{fig:mark_pipeline}.
	Encoding the NL sentence is relatively straightforward, while encoding the table is non-trivial since it exhibits underlying structures.
	In practise, we linearize the table into a flatten sequence so that it can be fed directly into the model.
	By inserting several special tokens to indicate the table boundaries, a linearized table can be represented as $T=\texttt{\small [TABLE]},t_1,{\cdots},t_N,\texttt{\small [COLUMN]},c_1,{\cdots},c_N$\footnote{If the model requires to predict the value, we attach the values of $c_i$ behind $c_i$ and separate each value with symbol `\&'.}.
	Here \texttt{\small [TABLE]} and \texttt{\small [COLUMN]} are special tokens indicating the region of table headers and column names respectively.
	Finally, we concatenate the linearized database $T$ with the NL sentence $\mathbf{x}$ and feed them to the encoder.
	For multi-turn settings, we concatenate the dialogue history and current question in reverse chronological order.

	\subsection{Structure Mark}\label{sec:structure_mark}
	\tif{Structure information plays an important role in text-to-SQL}.
	Structure includes database schema, conversation context and their relationships.
	Recently, in the research line of prompt tuning, \citet{aghajanyan2021htlm, chen2021knowprompt} make a step by making prompts with additional marks (some special tokens) to encode various structure information.
	Inspired by their works, we explore the idea of encoding structure information by designing \tif{structure mark} and inserting them into the input to make model structure-aware for text-to-SQL.
	
	In fact, we could simply extract structure information from the input and leverage all these structure information to improve the model.
	Concretely, the structure information can be roughly organized into three types: 
	(1) \tif{schema property} to expand the semantic information of schema;
	(2) \tif{database structure} to aggregate the information from schema neighbors; 
	(3) \tif{discourse structure} to supply the conversation context in history question.
	We give a valid example in Figure~\ref{fig:mark_pipeline}, which shows how to serialize the database schema and insert structure mark into the input.

	\subsubsection{Schema Property} 
	In a vanilla formulation, the semantic representation of schema only relies on the surface name that easily leads to disambiguation.
	For example, we could not tell the tiny difference between \textsc{Country.Continent} and \textsc{Country.Region}.
	However, there exists some obvious and available schema property to enrich the semantic information of schema, which will improve the correctness of alignment between question and database and finally boost the text-to-SQL performance.
	The information includes (1) the internal schema information from the database schema, such as primary key or column type (\textsc{Int}, \textsc{String} or \textsc{Date}); (2) the name-based linking information between question and schema that serve as the prior of schema linking; (3) the value-based linking information that augment the column representation via leveraging the database content information.
	
	Overall, it is relatively simple to obtain the schema property.
	Specifically, for the internal schema information, we could derive them from the database definition.
	For the name-based linking information, we enumerate the n-gram of question and schema then examine if they are aligned.
	To fine-grained model the alignment, we also add a prefix (exact or partial) before the `match'.
	For example, as shown in \textsc{Input} of Figure~\ref{fig:mark_pipeline}, column \textsc{Player\_id} has partially overlapped with the token `Player', thus we attach \textsc{Partial\_Match} before the column \textsc{Player\_id}.
	In order to compute this alignment, we simply derive the schema-linking results using fuzzy string-match following~\cite{sun-etal-2018-semantic,guo-etal-2019-towards, wang-etal-2020-rat}.
	For the value-based linking information, we first normalize their data formation (\eg{ uniform the representation of date}) then match them with the token in question.
	After obtaining these three types of schema property information, we insert them as prefixes ahead of the schema.
	Let's take an example for explanation: `\tif{\underline{{Partial-Match \& Primary-Key \& Integer}} Ranking.Player\_id}'.
	In this example, \tif{Ranking.Player\_id} is the column.
	The prefix with underline are structure mark that express two schema properties: (1) \tif{Partial-Match} indicates that \tif{Ranking.Player\_id} partially alignment with the question; (2) \tif{Primary-Key} and \tif{Integer} are all column properties.
	
	\subsubsection{Database Structure} 
	The database structure could improve the representation of the schema via aggregating information from neighboring nodes.
	As shown in figure~\ref{fig:mark_pipeline}, the database structure includes (1) the affiliation relations between columns and tables (e.g., \textsc{Id} of \textsc{Matches}); (2) the foreign key relations between columns (e.g., \textsc{WinnerId} links to \textsc{PlayerId}); (3) the tables relations (e.g., \textsc{Matches} links to \textsc{Ranking}).
	
	For affiliation relations, we attach the columns with their affiliated table such as \textsc{Matches.Id}.
	For the other two relations, we suppose the tables relations already includes the information of foreign key relations.
	Thus, we only consider the tables relations here.
	We adopt the template \ti{`schema1 links to schema2'} and fill in table names. 
	Then concatenate these pairs together and put them into input.
	In preliminary experiments, we find that affiliation relations and the tables relations could improve the model further.
	As for affiliation relations, they regulate the generation of column with correct table (\eg{ \textsc{Ranking.Year} but \textsc{Year} comes from \textsc{Matches}}).
	As for tables relations, it would directly affect the prediction of tables in \textsc{From} clause. 

	\subsubsection{Discourse Structure} 
	The history SQL provides the mentioned schema and potential intent in current turn \citep{zhang-etal-2019-editing,cai-wan-2020-igsql}, which might benefit entity co-reference resolution and intent ellipsis recovery. 
	Based on this observation, we insert previous SQL in the input to improve the modeling of discourse in terms of entity and intent.

	\subsection{Constrained Decoding}\label{sec:constrained_decoding}
	\tif{Decoding well-structured SQL} 
	means that generated SQL not only obeys the grammar but also faithful to the database schema. 
	We notice that BART is already skilled at learning SQL grammar.
	However, it sometimes struggles in schema prediction.
	For example, BART might outputs \textsc{Nation} instead of \textsc{Citizenship}.
	To address this problem, we first construct the prefix-trie of database schema and then filter out the illegal token during the beam search (Figure~\ref{fig:mark_pipeline}).
	Note that the difference between \ourmodel and grammar-based parser is that we do not need to specify the concrete grammar. 
	Instead, \ourmodel learns grammar through training, which shows better generalization to complex SQL\footnote{\ourmodel learns this complex SQL template efficiently: \tif{SELECT A (SQL) OP SELECT B (SQL)} \cite{wang-etal-2020-dusql}. 
		But it's bothersome to design the grammar.}.
	Compared with PICARD~\cite{Scholak2021:PICARD}, it's more like \ti{lexicon} mode, and not specify the grammar constrain. The time complexity is $O(1)$.

	\subsection{SQL Completion}
	In our study, we found that the generated SQL statements often miss some \textsc{JOIN} components, 
	since they are often not explicitly mentioned in natural language questions.
	\tif{To make the SQL complete}, we need to find back the potential missing \textsc{JOIN} components based on the database schema.
	Concretely, we first construct a schema graph, where the nodes are tables or columns, and edges are schema relationships.
	Then we try to find the tables and columns that are located in the shortest path of the existing tables and columns in an incompleted SQL.
	Take the case in Figure~\ref{fig:mark_pipeline} for example.
	It's an incompleted SQL that does not mention table \textsc{Matches} and column \textsc{Winner\_id} in \textsc{FROM} clause.
	We infer these two schemas based on their neighbors: \textsc{Players} and \textsc{Ranking}.
	\textsc{Matches} is located on the path of these two tables.
	And \textsc{Winner\_id} is the primary key of \textsc{Matches}.

	\section{Experimental Setup}
	\subsection{Datasets and Evaluation Metrics}
	We compare \ourmodel with previous task-specific parsers  using seven popular text-to-SQL benchmark datasets. 
	The dataset statistics are shown in Table~\ref{tb:dataset_stats} (Appendix~\ref{sec:dataset_stats}).
	To systemically compare our unified parser with previous task-specific models, we divided the dataset into three group to study the effectiveness of \ourmodel: 
	(1) \textbf{Multi-Domain}\footnote{ \ourmodel still shows a powerful performance in an internal single-domain dataset. Note that for Spider, CoSQL, SparC and DuSQL, we conduct experiments on dev set since the test set is not publicly available.}: WikiSQL~\cite{zhongSeq2SQL2017} and TableQA~\cite{sun2020tableqa}; 
	(2) \textbf{Multi-Table}: Spider~\cite{yu-etal-2018-spider} and DuSQL~\cite{wang-etal-2020-dusql}; 
	(3) \textbf{Multi-Turn}:
	CoSQL \cite{yu-etal-2019-cosql}, SParC~\cite{yu-etal-2019-sparc} and Chase~\cite{guo-etal-2021-chase}.

	For WikiSQL and TableQA, we utilize logic
	form accuracy (LX) and execution accuracy (EX) as evaluation metrics following~\citet{zhongSeq2SQL2017}. 
	For Spider and DuSQL, we report exact set match accuracy (EM) following~\citet{yu-etal-2018-spider}.
	For SParC, CoSQL and Chase, we report question match accuracy (QM) and interaction match accuracy (IM) following~\citet{yu-etal-2019-sparc}. 
	
	\begin{table*}[htb]
		\small
		\centering
		\begin{tabular}{ccccccccc}
			\toprule
			\multirow{3}{*}{\textbf{Model}}
			&  \multicolumn{4}{c}{\textbf{WikiSQL}} &  \multicolumn{4}{c}{\textbf{TableQA}} \\
			& \multicolumn{2}{c}{\textbf{Dev}} &  \multicolumn{2}{c}{\textbf{Test}} & \multicolumn{2}{c}{\textbf{Dev}} &  \multicolumn{2}{c}{\textbf{Test}} \\
			\cmidrule(r){2-3}  \cmidrule(r){4-5}  \cmidrule(r){6-7} \cmidrule(r){8-9}
			& \textbf{LX} & \textbf{EX} & \textbf{LX} & \textbf{EX} & \textbf{LX} & \textbf{EX} & \textbf{LX} & \textbf{EX}  \\
			\midrule
			\textit{Invasive Approaches} \\
			SQLNet \cite{xu2017sqlnet} & - & 69.8 & - & 68.0 & - & - & 61.4 & 67.2 \\
			Coarse2Fine \cite{dong-lapata-2018-coarse} & 72.5 & 79.0 & 71.7 & 78.5 & - & -  & 72.6 & 76.7\\
			SQLova \cite{sqlova} & 81.6 & 87.2 & 80.7 & 86.2 & - & -  & 81.7 & 85.8\\
			X-SQL \cite{He2019XSQLRS}& 83.8 & 89.5 & 83.3 & 88.7 & - & -  & 83.3 & 87.6\\
			F-SQL \cite{fsql} & - & - & 85.6 & \textbf{91.4} & - & - & 90.4 & 93.2 \\
			HydraNet \cite{Lyu2020HybridRN} & 83.6 & 89.1 & 83.8 & 89.2 & - & -  & - & -\\
			BRIDGE \cite{lin-etal-2020-bridging} & 86.2 & \textbf{91.7} & 85.7 & 91.1 & - & - & - & -\\
			\midrule
			\textit{Non-invasive Approaches} \\
			BART-Large \cite{lewis-etal-2020-bart} & 83.7 & 89.4 & 82.8 & 88.8 & 88.7 & 91.8 & 90.7 & 94.4\\
			SeaD \cite{sead}& 84.9 & 90.2 & 84.7 & 90.1 & - & - & - & -\\
			\ourmodel& \textbf{86.7} & \textbf{91.7} & \textbf{85.8} & \textbf{91.4} & \textbf{89.9} & \textbf{92.1} & \textbf{91.8} & \textbf{95.1}\\
			\bottomrule
		\end{tabular}
		\caption{
			Logical Form Accuracy (LX) and Execution Accuracy (EX) of the multi-domain setting.
			Note that we report the models without using Execution-Guided Decoding.
		}
		\label{tab:single-table}
	\end{table*}

	\begin{table}[htb]
		\small
		\centering
		\begin{tabular}{ccc}
			\toprule
			\textbf{Model} &  \textbf{Spider} & \textbf{DuSQL} \\
			\midrule
			\textit{Invasive Approaches} \\
			RYANSQL \cite{ryansql} & 66.6 & - \\
			IRNet  \cite{guo-etal-2019-towards} & 63.9 & 38.4 \\
			IRNetExt \cite{wang-etal-2020-dusql} & - & 59.8 \\
			RAT-SQL \cite{wang-etal-2020-rat}   & 69.7 & - \\
			BRIDGE \cite{lin-etal-2020-bridging} & 70.0 & - \\
			\midrule
			\textit{Non-invasive Approaches} \\
			T5-Base \cite{shaw-etal-2021-compositional}  & 57.1 & -  \\
			T5-Large \cite{Scholak2021:PICARD} & 65.3 & - \\
			$\text{T5-Large}^\dagger$ \cite{Scholak2021:PICARD} & 69.1 & - \\
			BART-Large \cite{lewis-etal-2020-bart}  & 64.5 & 82.2  \\
			\ourmodel & \textbf{70.0} & \textbf{84.3} \\
			\bottomrule
		\end{tabular}
		\caption{Exact-set-match accuracy (EM) of the multi-table setting.
			The PLM with $\dagger$ indicates the usage of constrained decoding~\cite{Scholak2021:PICARD}.
		}
		\label{tab:multi-table}
	\end{table}
	
	\begin{table}[htb]
		\small
		\centering
		\begin{tabular}{ccccccc}
			\toprule
			\multirow{2}{*}{\textbf{Model}}
			&  \multicolumn{2}{c}{\textbf{SParC}} &  \multicolumn{2}{c}{\textbf{CoSQL} }& \multicolumn{2}{c}{\textbf{Chase}}\\
			\cmidrule(r){2-3} \cmidrule(r){4-5} \cmidrule(r){6-7}
			& \textbf{QM} & \textbf{IM} & \textbf{QM} & \textbf{IM}  & \textbf{QM} & \textbf{IM}   \\
			\midrule
			\textit{Invasive} \\
			EditSQL  & 47.2 & 29.5 & 39.9  & 12.3  &37.7 &17.4 \\
			IGSQL & 50.7 & 32.5 & 44.1  & 15.8  & 41.4 &20.0 \\
			RAT-SQL  & 60.1 & 38.6 & 50.8  & 20.1  &35.1 &14.6 \\
			\midrule
			\textit{Non-invasive} \\
			BART & 55.0 & 36.5 & 47.1  &18.8  &37.6 &19.5 \\
			\ourmodel & \textbf{60.4} & \textbf{40.8} & \textbf{51.8} & \textbf{21.3} & \textbf{42.2} & \textbf{22.3} \\
			\bottomrule
		\end{tabular}
		\caption{Question Match (QM) and Interaction Match (IM) of the multi-turn setting. 
		}
		\label{tab:multi-turn}
	\end{table}

	\subsection{Baselines}
	For each setting, we select representative invasive models and non-invasive models as our baselines. For non-invasive methods, we only list the PLM in moderate-size to make fair comparison with \ourmodel under different settings. 
	
	\paragraph{Multi-Domain} 
	(1) SQLNet~\cite{xu2017sqlnet} is a sketch-based method. 
	(2) SQLova~\cite{sqlova} is a sketch-based method which integrates the pre-trained language model;
	(3) Coarse2Fine~\cite{dong-lapata-2018-coarse} first generates the SQL template then fills the value.
	(4) X-SQL~\cite{He2019XSQLRS} enhances the structural schema representation with the contextual embedding.
	(5) F-SQL~\cite{fsql} improves the representation of schema with table content.
	(6) HydraNet~\cite{Lyu2020HybridRN} uses column-wise ranking and decoding.
	(7) BRIDGE~\cite{lin-etal-2020-bridging} further leverages the database content to augment the column representation. 
	(8) SeaD~\cite{sead} trains a seq2seq model with schema-aware denoising objectives.
	
	\paragraph{Multi-Table} 
	(1) RYANSQL~\cite{ryansql} recursively predicts nested queries with sketch-based slot filling algorithm; 
	(2) IRNet~\cite{guo-etal-2019-towards} utilizes SemQL as an abstraction representation of SQL queries;
	(3) RAT-SQL~\cite{wang-etal-2020-rat} utilizes a complete relational graph attention neural network to handle various pre-defined relations;
	(4) IRNetExt~\cite{wang-etal-2020-dusql} extends IRNet to parse calculation questions and predict values.
	
	\paragraph{Multi-Turn}
	(1) EditSQL~\cite{zhang-etal-2019-editing} adopts an editing-based encoder-decoder model;
	(2) IGSQL~\cite{cai-wan-2020-igsql} proposes schema interaction graph encoder to utilize historicalal information of database schema items;
	(3) RAT-SQL-con~\cite{wang-etal-2020-rat} is the extension of RAT-SQL for multi-turn settings.
	
	\subsection{Implementation Details}
	We conduct the experiments using the Fairseq~\cite{ott-etal-2019-fairseq} to do data pre-processing, training and inference. The training process takes about 10 hours with four V100-16G GPUs.
	The hyper-parameters is in Appendix~\ref{sec:hyper_parameter}.
	Implementation of constrained decoding is based on~\citet{decao2020autoregressive,decao2020multilingual}.

	\section{Results and Analysis}
	\subsection{Main Results}
	The results of multi-domain, multi-table and multi-turn are listed in Table~\ref{tab:single-table}, \ref{tab:multi-table} and \ref{tab:multi-turn}.
	We run each \ourmodel experiment three times with different random seeds and report the mean. 
	Most results of previous models are reported by cited papers respectively. 
	For WikiSQL, we re-implement RAT-SQL with BERT-Large.
	For multi-turn settings, the results of invasive model are reported by \citet{guo-etal-2021-chase}.
Basically, \ourmodel achieves the excellent performance under all settings, which demonstrates the effectiveness of three extensions.

\begin{table}[htb]
	\small
	\centering
	\begin{tabular}{ccccc}
		\toprule
		\textbf{Model} & \textbf{\#Params
		} & \textbf{Spider} & \textbf{CoSQL}  & \textbf{SParC} \\
		\midrule
		T5-Base & 220M & 57.1 & - & -  \\
		$\text{T5-Base}^\dagger$ & 220M & 65.8 & - & -  \\
		T5-Large & 770M &  65.3 & - & -   \\
		$\text{T5-Large}^\dagger$ & 770M &  69.1 & - & -   \\
		T5-3B & 3B &  69.9 & 53.8 & -   \\
		$\text{T5-3B}^\dagger$ & 3B &  74.1 & 56.9 & -   \\
		BART-Large & 400M & 64.5 & 47.1 & 55.0  \\
		\midrule
		$\text{T5-Large}^\dagger$ + SM & 770M &  70.2 & - & -   \\
		\ourmodel & 400M & 70.0 & - & -  \\

		\bottomrule
	\end{tabular}
	\caption{Logical Form Accuracy (LX) of using different sizes of PLM for text-to-SQL tasks.
		The PLM with $\dagger$ indicates the usage of constrained decoding~\cite{Scholak2021:PICARD}. SM stands for structure mark.}
	\label{tab:compare_with_other_PLM}
\end{table}

\begin{table}[htb]
	\small
	\centering
	\begin{tabular}{cccc}
		\toprule
		\textbf{Model} &  \textbf{Spider} & \textbf{CoSQL}  & \textbf{SParC} \\
		\midrule
		\textbf{\ourmodel} & \textbf{70.0} & \textbf{51.8} & \textbf{60.4}  \\
		- SM  & 66.9(-3.1\%) & 48.7(-3.1\%) & 57.4(-3.0\%)  \\
		- CD & 67.5(-2.5\%) & 50.1(-1.7\%) & 57.8(-2.6\%)  \\
		- SC & 68.8(-1.2\%) & 50.8(-1.0\%) & 58.9(-1.5\%)  \\
		\bottomrule
	\end{tabular}
	\caption{The ablation study results of Logical Form Accuracy on \ourmodel.
		SM stands for structure mark, CD stands for constrained decoding, and SC stands for SQL completion.}
	\label{tab:ablation_study}
\end{table}

We also compare \ourmodel with other existing PLMs (Table~\ref{tab:compare_with_other_PLM}).
It can be observed that \ourmodel outperforms most of these PLMs even though they have much more parameters. 
Note that \ourmodel even slightly surpasses T5-3B (8x larger) in Spider.
And T5-large \citet{Scholak2021:PICARD} could be further improved by Structure Mark.

\subsection{Ablation Study}
To study the effect of our non-invasive approaches: structure mark, constrained decoding and SQL completion, we conduct ablation study on Spider, CoSQL and SParC.
As show in Table~\ref{tab:ablation_study}, the structure mark significantly boosts the performance of BART, implying that it does effectively represent 
the structure knowledge for autoregressvie language models.
At the same time, from the view of decoding, the constrained generation solution also boosts the performance by a large margin, since it alleviates the problem of generating invalid SQL.
As a complement to the above two extensions, SQL completion further improves the performance by inferring underlying tables in post-processing. 

\begin{table}[!htb]
	\small
	\centering
	\begin{tabular}{ccccc}
		\toprule
		\multirow{2}{*}{\textbf{Model}}
		& \multirow{2}{*}{\textbf{Setting}}  &   \multicolumn{2}{c}{\textbf{WikiSQL}} & \multicolumn{1}{c}{\textbf{Spider}}   \\ 
		&  & \textbf{Dev} & \textbf{Test} & \textbf{Dev} \\
		\midrule
		\multirow{3}{*}{RAT-SQL} 
		& single & 78.0 & 78.2 & 69.7 \\
		& joint & 77.5 & 77.7 & 68.8 \\
		& - & {\color{red}-0.5} &  {\color{red}-0.5} &  {\color{red}-0.9}\\
		\midrule
		
		\multirow{3}{*}{BART} 
		& single & 83.7 & 82.8 & 64.5 \\
		& joint & 83.4 & 83.7 & 63.9 \\
		& - & {\color{red}-0.3} &  {\color{mydarkgreen}+0.9} &  {\color{red}-0.6}\\
		
		\midrule
		\multirow{3}{*}{\ourmodel} 
		& single & 86.7 & 85.8 & 70.0 \\
		& joint & 86.5 & 86.8 & 68.9 \\
		& - & {\color{red}-0.2} &  {\color{mydarkgreen}+1.0} &  {\color{red}-1.1}\\
		
		\bottomrule
	\end{tabular}
	\caption{Logical Form Accuracy (LX) of joint training with WikiSQL (single-table) and Spider (multi-table). Note that RAT-SQL is trained with BERT-Large.}
	\label{tab:joint_table}
\end{table}

\begin{table}[!htb]
	\small
	\centering
	\begin{tabular}{ccccc}
		\toprule
		\textbf{Model} & \textbf{Setting} & \textbf{Spider} & \textbf{CoSQL}  & \textbf{SParC} \\
		\midrule
		\multirow{3}{*}{RAT-SQL} 
		& single & 69.7 & 50.8 & 60.1\\
		& joint & 68.2 & 50.0 & 58.2\\
		& - & {\color{red}-1.5} &  {\color{red}-0.8} &  {\color{red}-1.9}\\
		\midrule
		
		\multirow{3}{*}{BART} 
		& single & 64.5 & 47.1 & 55.0\\
		& joint & 67.5 & 50.1 & 57.8\\
		& - & {\color{mydarkgreen}+3.0} &  {\color{mydarkgreen}+3.9} &  {\color{mydarkgreen}+2.8}\\
		
		\midrule
		\multirow{3}{*}{\ourmodel} 
		& single & 70.0 & 51.8 & 60.4\\
		& joint & 70.8 & 52.9 & 60.9\\
		& - & {\color{mydarkgreen}+0.8} &  {\color{mydarkgreen}+1.1} &  {\color{mydarkgreen}+0.5}\\
		
		\bottomrule
	\end{tabular}
	\caption{Logical Form Accuracy (LX) of joint training across single-turn (Spider) and multi-turn (CoSQL and SParC) datasets. Note that RAT-SQL is trained with BERT-Large.}
	\label{tab:joint_turn}
\end{table}

\begin{figure*}[htb]
	\centering
	\includegraphics[width=1\linewidth]{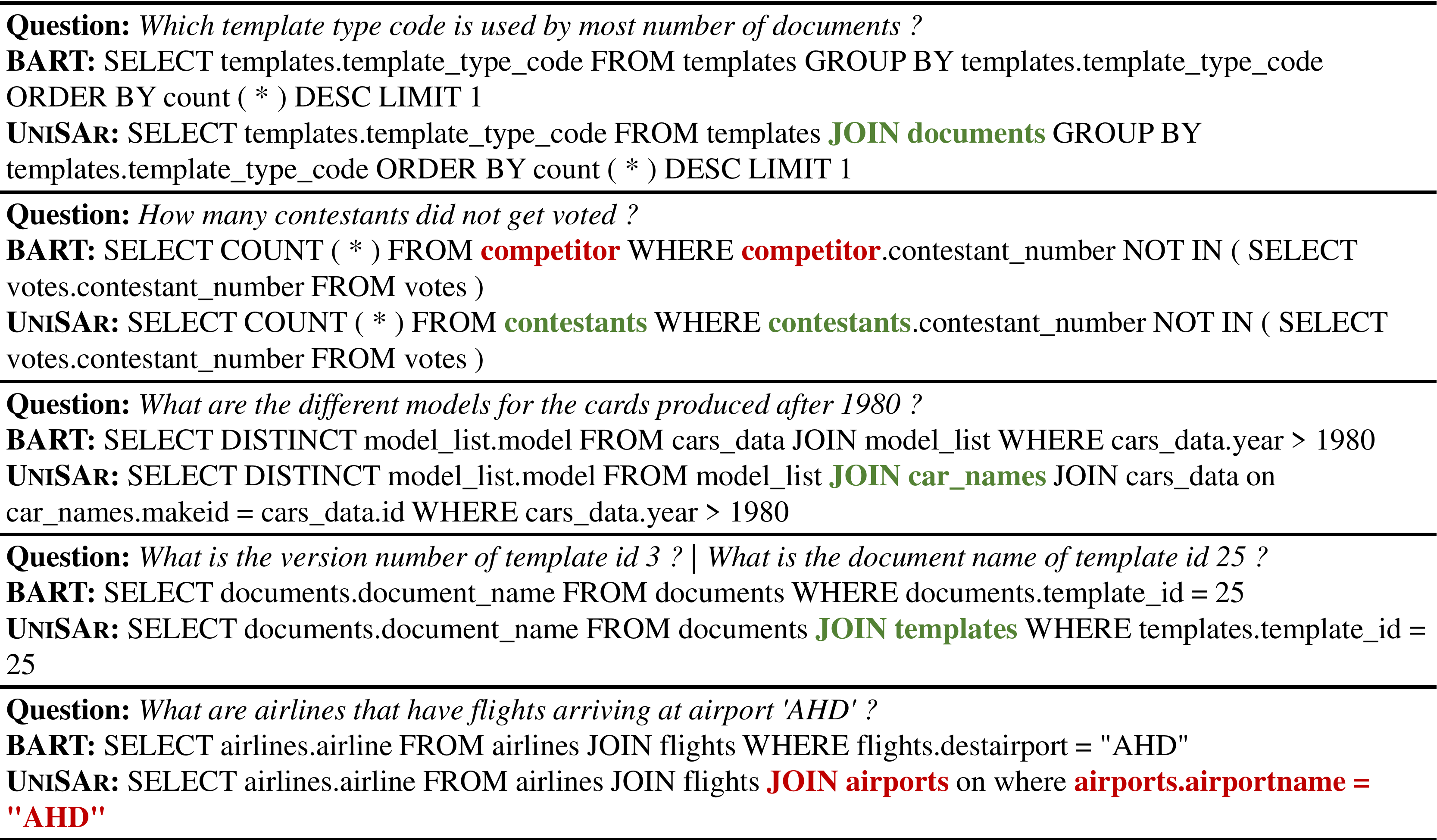}
	\caption{
		Case study: the first four cases are positive samples while the last one is negative.
		\texttt{FROM} conditions are omitted here for brevity.
	}
	\label{fig:case_study}
\end{figure*}

\subsection{Effect of Oracle Structure Mark}

To explore the upper-bound of performance boost that structure mark brings, we conduct experiments under oracle setting. 
Concretely, we derive oracle schema-linking for Spider by human annotation from \citet{lei-etal-2020-examining} and oracle previous SQL for CoSQL, SParC and Chase.
Experimental results show that Spider gets a 1.2\% performance boost.
And for these multi-turn datasets (SParC, CoSQL, Chase), they receive 5.9\%, 8.7\%, 11.7\% boost respectively.
It indicates the trends that \ourmodel will improve further if we could obtain higher-quality structure mark.

\subsection{Generalization Across Different Datasets}\label{sec:generation}
The simple architecture of \ourmodel enables different tasks to share the sample training protocol. It's easy to perform multi-task train.
To examine this,
we conduct two types of multi-task: (a) single-table + multi-table and (b) single-turn + multi-turn. 
For single-table and multi-table, we train WikiSQL and Spider jointly (Table~\ref{tab:joint_table}). 
It can be seen that all these three models degrade.
But BART and \ourmodel generally show robustness to the unbalanced distribution of data amount (80k vs. 10k) and SQL format (simple vs. complex).
For single-turn and multi-turn, we train Spider, CoSQL and SParC jointly (Table~\ref{tab:joint_turn}). 
It indicates that both BART and \ourmodel show a positive trend by jointly training, whereas RAT-SQL not.
The divergence could be attributed to the task-specific modules (\eg grammar-based decoder) which make the model easily overfit to specific dataset. 

\subsection{Case Studies}
In Figure~\ref{fig:case_study}, we compare the SQL generated by our model \ourmodel with those created by the vanilla BART.
We notice that \ourmodel performs better than the BART, especially on examples that involve the \texttt{JOIN} operation of multiple tables.
For example, in the first case where BART fails to identify the existence of table \textsc{Documents}.
For comparison, \ourmodel successfully predicts the connection of two tables since the structure mark presents the database structure.
In the second case, vanilla BART predicts a token (e.g., \textsc{Competitor}) that does not exist in the database schema. This will cause an ill-formed SQL but \ourmodel ensures the faithful generation with constrained decoding. 
Moreover, in the third case, we can find that SQL completion infers the missing table \textsc{Car\_names} based on the matched table \textsc{Model\_list} and \textsc{Cars\_data}.
In the fourth case from multi-turn setting, \ourmodel still outperforms the BART in contextual modeling with effectively encoding the information of dialogue history.
However, in the last case, our \ourmodel is awkward to predict unnecessary table \textsc{Airports}.
This error perhaps can be attributed to inappropriate structure mark due to inaccurate schema-linking. 
A high precision structure mark will alleviate this problem.

\section{Conclusion and Future Work}
In this paper, we propose \ourmodel, a simple-yet-effective model to solve text-to-SQL across various settings in a unified framework.
Concretely, we simply extend BART, an off-the-shelf autoagressive language model, with three non-invasive extensions to make it structure-aware.
Experimental results demonstrate the effectiveness of our \ourmodel on seven well-known text-to-SQL datasets.
\ourmodel achieves highly comparable or better performance to the most advanced specifically-designed text-to-SQL models.
\ourmodel also shows excellent generalizability in joint training and achieves overall improvements. 

With the rapid development of the PLM model, we believe that incorporating structural knowledge from data will be a tendency (data-centric). Structure mark would also benefit the tasks that require modeling the structure knowledge in input data. 

\bibliography{acl}
\bibliographystyle{acl_natbib}

\clearpage
\appendix
\section{Datasets Statistics}
\label{sec:dataset_stats}

Table~\ref{tb:dataset_stats} lists the dataset we used.
We divided the dataset into three group to study the effectiveness of \ourmodel: 
(1) \textbf{Multi-Domain}: WikiSQL~\cite{zhongSeq2SQL2017} and TableQA~\cite{sun2020tableqa}; 
(2) \textbf{Multi-Table}: Spider~\cite{yu-etal-2018-spider} and DuSQL~\cite{wang-etal-2020-dusql}; 
(3) \textbf{Multi-Turn}:
CoSQL \cite{yu-etal-2019-cosql}, SParC~\cite{yu-etal-2019-sparc} and Chase~\cite{guo-etal-2021-chase}.

\begin{table*}[!ht]
	\centering
	\scalebox{0.9}{
		\begin{tabular}{c|cccccccc}
			\hline
			Dataset & \# Lan.  & \# Ques. & \# SQL Len. & \# Table & \!\!\# DB \!\! & \# MD & \# MTa & \# MTu\!\!\!\\ \hline
			WikiSQL \cite{zhongSeq2SQL2017} & English  & 80,654 & 10.6 & 26,521 & 26,521 & \checkmark & \XSolidBrush & \XSolidBrush \\
			TableQA  \cite{sun2020tableqa} & Chinese   & 49,974   & 9.1 & 5,291 & 5,291 & \checkmark & \XSolidBrush & \XSolidBrush \\
			Spider \cite{yu-etal-2018-spider} &  English  & 10,181  & 21.7 & 1,020 & 200  & \checkmark & \checkmark & \XSolidBrush\\
			DuSQL \cite{wang-etal-2020-dusql} &  Chinese & 23,797  & 20.2 & 820 & 200 & \checkmark & \checkmark & \XSolidBrush \\
			CoSQL \cite{yu-etal-2019-cosql} &  English & 15,598   & 18.4 & 1,020 & 200 & \checkmark & \checkmark & \checkmark  \\
			SParC \cite{yu-etal-2019-sparc} &  English & 12,726  & 17.8 & 1,020 & 200 & \checkmark & \checkmark &\checkmark   \\
			Chase \cite{guo-etal-2021-chase}  & Chinese  & 17,940  & 20.9 & 1,020 & 200 & \checkmark &\checkmark &\checkmark \\ \hline
	\end{tabular}}
	\caption{Comparisons of seven text-to-SQL datasets.
		MD (multi-domain), MTa (multi-table) and MTu (multi-turn) denote the research focuses of these datsets and \# SQL Len. denotes the complexity of predicted SQL.}
	\label{tb:dataset_stats}
\end{table*}

\section{Hyper-parameters}
\label{sec:hyper_parameter}
We adopt the BART-Large and set the task of Fairseq as \textsc{translation\_from\_pretrained\_bart}.
The learning rate is $1e\text{-}5$. 
The max tokens is 1400 for V100-16G GPU.
We adopt the polynomial\_decay with 5,000 warmup-updates.
The dropout~\citep{Srivastava2014DropoutAS} rate is 0.1.
Optimizer is Adam~\citep{Kingma2015AdamAM} with the default parameters.
The max-update is set to 10,000.
Empirically, the model obtained best performance about 7000 steps (about 10 $\sim$ 15 epochs) in Spider, CoSQL and SParC. The Fairseq~\citep{ott-etal-2019-fairseq} dynamically tunes the batch size to realize higher GPU utilization.

\end{document}